\pdfoutput=1

\documentclass[11pt]{article}

\usepackage{ACL2023}

\usepackage{times}
\usepackage{latexsym}
\usepackage{subfiles}
\usepackage{amsmath}
\usepackage{amsfonts}
\usepackage{bm}
\usepackage{cite}
\usepackage{algorithm}
\usepackage{algpseudocode}
\usepackage{adjustbox}
\usepackage{booktabs}
\usepackage{tcolorbox}
\usepackage{multirow}
\usepackage{xspace}
\usepackage{xcolor,colortbl}
\usepackage{pifont}
\usepackage{graphicx} 
\usepackage{mdframed}
\usepackage{lipsum}
\usepackage{amssymb}
\usepackage{siunitx}

\definecolor{darkgreen}{rgb}{0.0, 0.5, 0.0}
\newcommand*{\method}{\textsc{OFCRE}\@\xspace}
\usepackage[T1]{fontenc}

\usepackage[utf8]{inputenc}

\usepackage{microtype}

\usepackage{inconsolata}

\title{Few-shot Continual Relation Extraction via Open Information Extraction}

\author{
\textbf{Thiem Nguyen\textsuperscript{1}\footnotemark[1]},
  \textbf{ Anh Nguyen\textsuperscript{2}\footnotemark[1]},
  \textbf{Quyen Tran\textsuperscript{3}\footnotemark[1]},\\
  \textbf{Tu Vu\textsuperscript{4}},
  \textbf{Diep Nguyen\textsuperscript{5}},
  \textbf{Linh Ngo\textsuperscript{1}\footnotemark[1]}, 
  \textbf{Thien Nguyen\textsuperscript{6}}
  \bigskip \\
\textsuperscript{1}Hanoi University of Science and Technology,
\textsuperscript{2}Oraichain Labs Inc., US,
\textsuperscript{3}VinAI Research
 \\
 \textsuperscript{4}Bytedance,
\textsuperscript{5}VNU University of Engineering and Technology,
\textsuperscript{6}University of Oregon,
}

\begin{document}
\maketitle

\renewcommand{\thefootnote}{\fnsymbol{footnote}}
\footnotetext[1]{Equally contributed.}
\footnotetext[2]{Corresponding author: \href{mailto:email@domain}{linhnv@soict.hust.edu.vn}}
\renewcommand*{\thefootnote}{\arabic{footnote}}

\begin{abstract}
Typically, \textit{Few-shot Continual Relation Extraction (FCRE)} models must balance retaining prior knowledge while adapting to new tasks with extremely limited data. However, real-world scenarios may also involve unseen or undetermined relations that existing methods still struggle to handle. To address these challenges, we propose a novel approach that leverages the \textit{Open Information Extraction} concept of \textit{Knowledge Graph Construction (KGC)}. Our method not only exposes models to all possible pairs of relations, including determined and undetermined labels not available in the training set, but also enriches model knowledge with diverse relation descriptions, thereby enhancing knowledge retention and adaptability in this challenging scenario. In the perspective of KGC, this is the first work explored in the setting of Continual Learning, allowing efficient expansion of the graph as the data evolves. 
Experimental results demonstrate our superior performance compared to other state-of-the-art FCRE baselines, as well as the efficiency in handling dynamic graph construction in this setting.

\end{abstract}

\section{Introduction}

Few-shot Continual Relation Extraction - FCRE \citep{qin-joty-2022-continual, chen-etal-2023-consistent} has emerged as a challenging problem, where models must continuously adapt to identify new relations in each sentence with a limited amount of data while preserving all the previous information accumulated over time. Current approaches have demonstrated researchers' efforts to simultaneously maintain important characteristics of FCRE models, including the (i) flexibility to adapt to new relations, (ii) preserve generalization, and (iii) effectively avoid forgetting. These requirements have been recently addressed through a special regularization strategy \citep{tran-etal-2024-preserving, DBLP:conf/acl/WangWH23}. Besides, they can also be achieved through memory enrichment strategies \citep{ma-etal-2024-making} or by designing special prompt inputs \citep{DBLP:conf/acl/ChenWS23} to guide the learning of models.

However, the \textit{existing work are somewhat impractical} when most of them focus solely on optimizing models in {the scenario with a predefined set of entities and relations}. Specifically, the models are only trained and tested on a given set of entities in each sentence, with the corresponding targets also being predefined, making the questions of the applicability of current approaches. Nevertheless, in the real open-world scenarios \citep{DBLP:conf/www/XuLSY19, Mazumder2024}, more potential pairs of entities can appear in a testing sample, which is often uncovered in the training dataset. To deal with this drawback, several methods \citep{WANG2023151, zhao-etal-2025-dynamic, zhao-etal-2023-open, meng-etal-2023-rapl} have considered unknown labels, but their training only relies on available information, including provided entities and relations from the training set, and poorly considers a NOTA (None Of The Above) label for all possible relations that are uncovered. 
Therefore, it is essential to develop FCRE models, which are able to
\textit{(a) identify relations between possible pairs of entities in a sentence by holistically taking advantage of the available information from training dataset, and, most importantly, (b) indicate whether the corresponding relations are known or not, and whether they are reasonable}. 

Related to these requirements, Open Information Extraction - OIE \citep{liu2022open, zhou2022survey,li2023evaluating} has been known as a solution for covering all possible pairs of entities and relations. Most recently, EDC \citep{zhang-soh-2024-extract} has proposed an efficient OIE strategy to extract any triples from given texts to construct relational graphs. 
However, this work and other related ones in the field of Knowledge Graph Construction (KGC) are only limited to exploiting information from a fixed dataset, while their ability in Fewshot Continual Learning scenarios, where there are always emerging relations, has not been explored. In this challenging scenario, the models need to adapt to information arriving sparsely and continuously over time. Besides, adapting the model to new information also poses challenges in reproducing relations from the previous knowledge that has been integrated into the graph.

To address the challenges of FCRE in practical settings and to explore the potential of KGC models in this complex scenario, we propose a novel solution inspired by the concept of Open Information Extraction (OIE). More specifically, our approach involves leveraging Named Entity Recognition (NER) \citep{zaratiana2023glinergeneralistmodelnamed} to extract and analyze all possible pairs of entities, focusing on identifying both determined and undetermined relations between them. We then utilize OIE for open extraction and generating corresponding descriptions that help effectively align sample representations. In addition, our sample-description matching schema is also a more effective solution for building KGC models in the latest SOTA \citep{zhang-soh-2024-extract}, which was based on description-description constraints. For testing, we dedicatedly employ OIE to filter out non-relational samples before they enter the FCRE modules, thereby specifically determining whether a tested sample has a known or unknown relation, or whether it is reasonable, thus improving overall testing performance. 

In summary, our key contributions include:
\begin{itemize}
\item We propose a novel solution for FCRE via Open Information Extraction, which 
effectively deals with unknown labels by considering all possible pairs of entities and corresponding relations in each sentence.
Based on this, we can determine whether a pair has a known or reasonable relationship, which existing methods have not considered.


\item For the first time, we consider the potential of Knowledge Graph Construction (KGC) models in the setting of Continuous Learning, where there are always emerging relations. In addition, our novel approach effectively elevates schema matching in KGC as minimizing errors from LLM-based schema extraction.

\item Experimental results indicate our superior performance over FCRE baselines in all cases with N/A relation or not. In addition, the superiority over KGC models in this extreme scenario is also demonstrated.




\end{itemize}

\section{Related Work and Background}
\subsection{Related Work}
Most existing FCRE methods \citep{DBLP:conf/acl/WangWH23, hu-etal-2022-improving, DBLP:conf/coling/MaHL024, tran-etal-2024-preserving} have utilized contrastive learning and memory replay techniques to significantly mitigate catastrophic forgetting. However, these approaches largely overlook the present of undetermined relations — relations that are unseen or nonexistent, which remains a critical gap in real-world applications. On the other hand, several methods \citep{WANG2023151, zhao-etal-2025-dynamic, zhao-etal-2023-open, meng-etal-2023-rapl} have considered unknown labels, but their training only relies on available information, including provided entities and relations from the training set, and poorly considers a NOTA (None Of The Above) label for all possible relations that are uncovered. 

Historically, relation extraction research has explored various types of undetermined relations. For example, prior work has defined “no relation (NA)” \citep{xie-etal-2021-revisiting} as sentences with no meaningful relationship between entities, “out-of-scope (OOS)” \citep{liu-etal-2023-novel} as relations outside predefined sets, and “none of the above (NOTA)” \citep{zhao-etal-2023-open} as relations that do not match any known type. While these studies address specific aspects of undetermined relations, their approaches are often simplistic and unrealistic, focusing on single labeled entity pairs rather than considering multiple possible relations within sentences.

Moreover, Open Information Extraction (OIE) has emerged as a powerful tool for open entity and relation extraction, particularly for knowledge graph construction, due to its ability to operate without predefined schemas. Recent studies \citep{li2023evaluating} highlight the strong performance of large language models (LLMs) in OIE tasks. For instance, EDC \citep{zhang-soh-2024-extract} propose an end-to-end pipeline that extracts, defines, and canonicalizes triplets to build knowledge graphs more efficiently. This pipeline includes three phases: (1) Open Information Extraction, where entity-relation triplets are freely extracted from text; (2) Schema Definition, where entity and relation types are defined based on extracted triplets; and (3) Schema Canonicalization, which standardizes relations to fit a target schema. This approach is particularly promising for handling undetermined relations, as it enables the extraction of relations beyond predefined sets.

\subsection{Background}
\subsubsection{Problem Definition}
Few-Shot Continual Relation Extraction (FCRE) requires a model to sequentially acquire new relational knowledge while retaining previously learned information. At each task $t$, the model is trained on a dataset $D^t = \{(x_i^t, y_i^t)\}_{i=1}^{N \times K}$, where $N$ denotes the number of labels provided in the set of relations $R^t$, and $K$ represents the limited number of training instances per relation (i.e., "$N$-way-$K$-shot" paradigm \citet{chen-etal-2023-consistent}). Each training example $(x, y)$ consists of a sentence $x$, which is originally given two entities $(e_h, e_t)$ and the associated relation labels $y \in R^t$. After completing task $t$, previously observed datasets $D^t$ are not extensively reused. The model's final evaluation is conducted on a test set comprising all encountered relations $\tilde{R}^T = \bigcup_{t=1}^{T} R^t$.

Beyond the standard setting and requirements of FCRE, in terms of mitigating forgetting and overfitting, our work aims at designing advanced models, which are capable of continuously capturing and recognizing new relational knowledge, which is not available in the training set.

\subsubsection{Latent Representation Encoding}
One of the fundamental challenges in relation extraction lies in effectively {encoding the latent representation} of input sentences, particularly given that Transformer-based models \citep{vaswani2017attention} produce structured matrix representations. In this study, we adopt an approach inspired by \citet{ma-etal-2024-making}. Given an input sentence $x$ that contains a head entity $e_h$ and a tail entity $e_t$, we transform it into a Cloze-style template $T(x)$ by inserting a \texttt{[MASK]} token to represent the missing relation. The structured template is defined as:

\begin{align}
\begin{aligned}
  T({x}) = \; &x \left[v_{0:n_0-1}\right] e_h \left[v_{n_0:n_1-1}\right] [\texttt{MASK}] \\
  &\left[v_{n_1:n_2-1}\right] e_t \left[v_{n_2:n_3-1}\right].
\label{eq:template}
\end{aligned}
\end{align}

where $[v_i]$ represents learnable continuous tokens, and $n_i$ denotes the respective token positions in the sentence. In our specific implementation, BERT’s \texttt{[UNUSED]} tokens are used for $[v]$. We set the soft prompt length to 3 tokens, with $n_0, n_1, n_2$, and $n_3$ assigned values of 3, 6, 9, and 12, respectively. The transformed input $T(x)$ is then processed through a pre-trained BERT model, encoding it into a sequence of continuous vectors. The hidden representation $z$ of the input is extracted at the position of the \texttt{[MASK]} token:

\begin{equation}
    z = \mathcal{M} \circ T(x)[\text{position}(\texttt{[MASK]})],
\end{equation}

where $\mathcal{M}$ represents the backbone language model. The extracted latent representation is subsequently passed through a multi-layer perceptron (MLP), allowing the model to infer the most appropriate relation for the \texttt{[MASK]} token.
\section{Proposed Method}


\begin{figure*}[t]
    \centering
    \includegraphics[width=\linewidth]{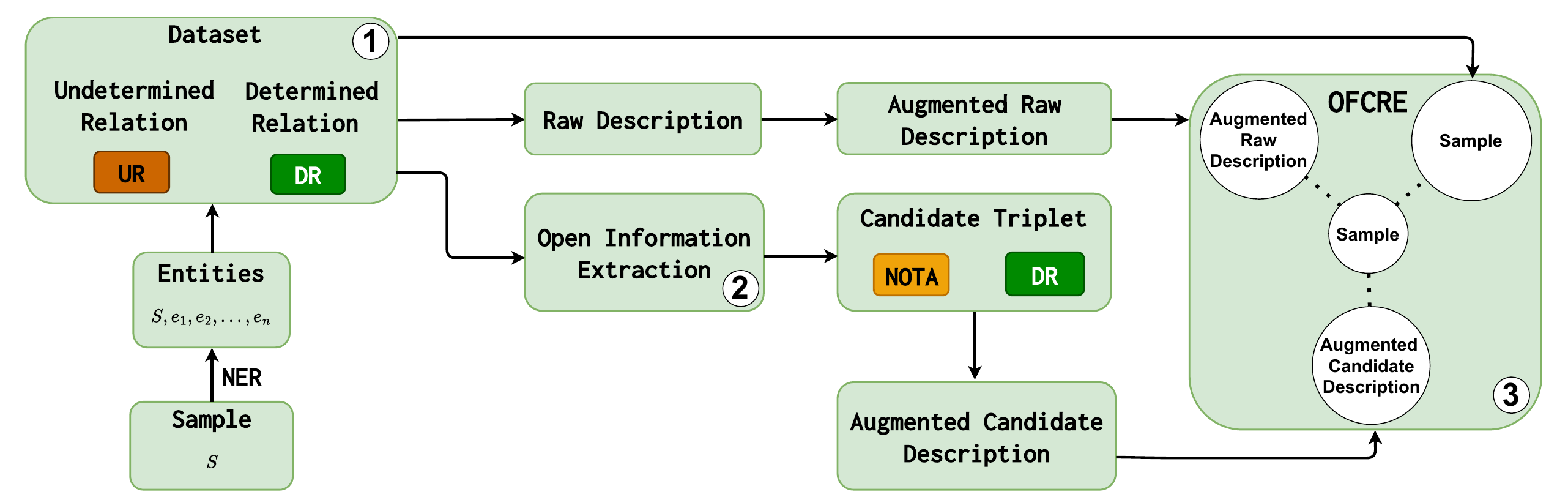}
    \caption{\textbf{{Our general framework:}} \textbf{\textit{(1) Open Dataset Construction}}, which creates training and testing datasets that account for undetermined relationships; \textbf{\textit{(2) Open Information Extraction}}, which is used to prepare candidate triplet (i.e., determined relationships of entity pairs); and  \textbf{\textit{(3) Training Open Few-shot Continual Relation Extraction (OFCRE)}} module, which aligns embeddings of sentences, original descriptions and candidate descriptions.}
    \label{fig:overview}
\end{figure*}





In real-world applications, Relation Extraction aims to identify relationships between all possible pairs of entities within documents. However, a significant challenge arises from the presence of \textit{Undetermined Relations} (UR) between entities, which is either \textit{not applicable} or \textit{unknown} as Appendix \ref{app:example}. Particularly, UR can be categorized into two types as follows:

\begin{itemize}
    \item \textbf{No Relation (NA)}: Used when no meaningful relationship exists between entities. 
    \item \textbf{None Of The Above (NOTA)}: Used when an entity pair does not fit any predefined relations.
\end{itemize}

Related to this problem, previous studies \citep{zhao-etal-2025-dynamic} primarily focus on NOTA relations and evaluate models using a simple threshold on the test set including unseen relations. However, this approach does not reflect real-world scenarios as they still only considered a predefined set of entity pairs and corresponding relations when training. 

This section will present our novel approach to dealing with this problem, going from extracting all possible entities to create an open dataset (Section \ref{sec:dataset}), to how OIE is utilized to support FCRE with undetermined relations (Section \ref{sec:oie}), and finally our training and inference procedures in Section \ref{sec:fcre}.  

\subsection{Open Dataset Construction}
\label{sec:dataset}

This is the first stage to extract all possible entities in a sentence for the training phase. Particularly, we employ a Named Entity Recognition (NER) model as Figure \ref{fig:overview}. However, extracted entities may not perfectly align with the original dataset annotations, thus we merge the extracted entities with overlapped ones in the benchmark dataset to ensure the consistency. 

Beyond this step, we assign labels to all possible entity pairs. If an extracted entity pair matches a predefined relation in the benchmark dataset, it is categorized as a \textit{determined relation} \textbf{(DR)}; otherwise, it is classified as an \textit{undetermined relation} \textbf{(UR)}. This approach results in a more comprehensive and realistic dataset, incorporating both original relations and \textit{undetermined relations} as newly labeled instances with descriptions. Each extracted entity pair with sample from the merged list is treated as an independent instance, rather than just a sample with the original entity pair, serving as input for the relation extraction task. Consequently, the dataset size significantly increases due to the large number of \textit{undetermined relations}, making it more reflective of real-world scenarios.





\subsection{Open Information Extraction}
\label{sec:oie}


Unlike existing FCRE methods, this module in Fig.\ref{fig:overview} aims to identify unseen relations, thereby expanding the scope of knowledge extraction for more efficient training. In particular, we employ the OIE module of EDC to extract relations between entities without any predefined label set. To this end, we employ ChatGPT-4o-mini to generate candidate triplets that contain relations and follow a structured prompting approach, as illustrated in Fig.\ref{tab:relation_description_prompt1}.


\subsection{FCRE via OIE (OFCRE)} 
\label{sec:fcre}
This section in Fig.\ref{fig:overview} presents our training and testing process. Paticularly, we demonstrate how expanding the relation set with UR aids efficient training and enables the model to handle unseen labels.

\subsubsection{Training phase}

\paragraph{Data Augmentation}
\label{sec:augment}

Overall, relation descriptions from both training datasets and LLM generation are typically concise, generic, and applicable to multiple samples \citep{han-etal-2018-fewrel, zhang-etal-2017-position}. However, relying solely on these limited descriptions can constrain model performance, motivating us to enhance them with greater diversity.
\begin{itemize}
    \item For each original description, we augment it with $K$ additional samples. Each sample includes an example sentence closely related to the target relation, thereby improving the alignment between the embeddings of the relation and the corresponding diverse descriptions.

 \item Similarly, to utilize the candidate triplet produced by the OIE module, an additional prompt is crafted to deliver $K$ distinct \textit{candidate relation descriptions} with examples. This aids in examine the surrounding context to formulate a \textit{candidate description} tailored to the identified relation. These context-sensitive descriptions serve as enhanced refinements of the original ones, offering more accurate and detailed representations.
\end{itemize}


 These enriched descriptions contribute to better model generalization. Note that description augmentation is applied only to seen relation types, not to undetermined relations. Further details on both types of descriptions can be found in Appendix \ref{app:prompt}.

\paragraph{Objective Functions}
\label{sec:loss}
\paragraph{\textit{Hard Soft Margin Loss (HSMT)}} 
To enhance the distinction between different relations, HSMT integrates both hard and soft margin triplet loss principles \citep{hermans2017defense}, dealing with the most challenging positive and negative samples while maintaining flexibility through a soft margin. Formally, the loss function is defined as:

\begin{multline}
\mathcal{L}_{\textrm{HSMT}}({x}) = 
- \log \bigg(1 + \max_{{p} \in P(\bm{x})} e^{\xi(\bm{z}_x, \bm{z}_p)} \\
- \min_{{n} \in N({x})} e^{\xi(\bm{z}_x, \bm{z}_n)}
\bigg),
\end{multline}

where $\xi(\cdot, \cdot)$ denotes the Euclidean distance function. This formulation effectively maximizes the separation between the hardest positive and hardest negative samples while allowing for adaptive margin flexibility, improving representations in the latent space.

\paragraph{\textit{Weighted Mutual Information Loss}}
This loss aims to maximize the mutual information between relation embedding $\bm{z}_i$ and its corresponding retrieved description embedding $\bm{d}_i$ of sample $\bm{x}_i$, ensuring a more informative alignment. Following \citet{DBLP:journals/corr/abs-1807-03748}, the mutual information $MI(x)$ between $\bm{z}_i$ and its corresponding label description satisfies:


\begin{equation}
    MI(x) \geq \log B + \textnormal{InfoNCE} (\{ x_i\}_{i=1}^B; h),
\end{equation}

\begin{multline}
\textnormal{where InfoNCE} (\{ x_i\}_{i=1}^B; h) = \\
\frac{1}{B} \sum_{i=1}^B \log \frac{\sum_{k=1}^K h(\bm{z}_i, \bm{d}_i^k)}{\sum_{j=1}^B \sum_{k=1}^K h(\bm{z}_j, \bm{d}_j^k)},
\end{multline}

and

\begin{equation}
h(\bm{z}_j, \bm{d}_j^k) = \exp \left(\frac{\bm{z}_j^T W \bm{d}_j^k}{\tau} \right).
\end{equation}

Here, $\tau$ is the temperature parameter, $B$ is the mini-batch size, and $W$ is a trainable weight matrix. We define $P(x)$ as the set of positive samples (same-label pairs) and $N(x)$ as the set of negative samples (different-label pairs). 

Given the imbalance caused by a high proportion of \textit{Undetermined Relation (UR)} labels, we introduce a weight adjustment based on the number of samples for each unique relation type in the batch:

\begin{equation}
    w_{x} = \frac{B}{\|P(x)\| + \sum_{y \in N(x)} \frac{\|P(x)\|}{\|P(y)\|}}
\end{equation}

The final Weighted MI loss function is formulated as:

\begin{multline}
\mathcal{L}_{\textrm{WMI}}(x, d) = 
- w_{x}\log \frac{ \sum_{k=1}^K
    h(\bm{z}_x, \bm{d}_x^k)
}{Z(x,d)}
\end{multline}

\noindent where
\begin{equation}
Z(x,d) = \sum_{k=1}^K h(\bm{z}_x, \bm{d}_x^k) + \sum_{n \in N(x)} \sum_{k=1}^K h(\bm{z}_x, \bm{d}_n^k)
\end{equation}

This loss is applied not only to the \textit{raw description} but also to the \textit{candidate description} to enhance the learning of sample representations. These types of descriptions serve as stable reference points when learning known relation types within batches containing numerous undetermined relations. We optimize this description loss as follows:

\begin{equation}
\mathcal{L}_{\textrm{Des}} = \mathcal{L}_{\textrm{WMI}}(x, d) + \mathcal{L}_{\textrm{WMI}}(x, c) 
\end{equation}

\paragraph{\textit{Training Objective Function}}
The final optimization objective combines losses that align both \textit{sample-to-sample} and \textit{sample-to-description} representations, incorporating weighted coefficients:

\begin{equation}
\begin{aligned}
\mathcal{L}(x) &= \mathcal{L}_{\textrm{Samp}} + \mathcal{L}_{\textrm{Des}}  \\ 
               &= \alpha_{x} \mathcal{L}_{\textrm{HSMT}}(x) 
               + \alpha_{xd} \mathcal{L}_{\textrm{WMI}}(x,d) \\
               &\quad + \alpha_{xc} \mathcal{L}_{\textrm{WMI}}(x,c)
\end{aligned}
\label{eq:total_loss}
\end{equation}

where $\alpha_{x}$, $\alpha_{xd}$ and $\alpha_{xc}$ are tunable hyperparameters controlling the relative contribution of each loss term.
\label{sec:procedure}
\paragraph{Training Procedure:}
Algorithm \ref{alg:training} provides a structured approach for training at each task $\mathcal{T}^j$. Here, $\Phi_{j-1}$ represents the model state after learning from the previous $j-1$ tasks. Following a memory-based continual learning strategy, we maintain a memory buffer $\tilde{M}_{j-1}$, which retains selected representative instances from earlier tasks ${\mathcal{T}^1, \dots, \mathcal{T}^{j-1}}$. Additionally, we keep track of a relation description set $\tilde{E}_{j-1}$ and a candidate description set $\tilde{C}_{j-1}$, which store descriptions of previously encountered relations.

\begin{enumerate}
    \item \textbf{Initialization} (Lines 1--2):  
    The model parameters for the current task, $\Phi_j$, are inherited from $\Phi_{j-1}$. The relation description sets $\tilde{E}_j$ and $\tilde{C}_j$ are then updated by integrating new relation details from $E_j$ and $C_j$, respectively.
    
    \item \textbf{Task-Specific Training} (Line 3):  
    To accommodate new relations introduced in $\mathcal{T}^j$, $\Phi_j$ is trained on $D_j$.
    
    \item \textbf{Memory Management} (Lines 4--8):  
    For each relation $r \in R_j$, we choose $L$ key samples from $D_j$ that are closest to the 1-means centroid of the relation class. These selected samples form memory components $M_r$, contributing to the refined memory set $\tilde{M}_j = \tilde{M}_{j-1} \cup M_j$, alongside the expanded relation set $\tilde{R}_j = \tilde{R}_{j-1} \cup R_j$.

    \item \textbf{Prototype Construction} (Line 9):  
    A prototype set $\tilde{P}_j$ is generated based on the updated memory $\tilde{M}_j$ for inference purposes.

    \item \textbf{Memory-Based Training} (Line 10):  
    The model $\Phi_j$ is further refined by training on the enhanced memory dataset $\tilde{M}_j^*$ to reinforce its ability to retain and recall previously learned relations.
    

\end{enumerate}

\begin{algorithm}
\caption{Training procedure at each task $\mathcal{T}^j$}
\label{alg:training}
\begin{algorithmic}[1]
\Require $\Phi_{j-1}$, $\tilde{R}_{j-1}$,  $\tilde{M}_{j-1}$, $\tilde{E}_{j-1}$, $\tilde{C}_{j-1}$, $D_j$, $R_j$, $E_j, C_j$ 
\Ensure $\Phi_j, \tilde{M}_j, \tilde{K}_j, \tilde{P}_j$
\State Initialize $\Phi_j$ from $\Phi_{j-1}$
\State $\tilde{E}_j \gets \tilde{E}_{j-1} \cup E_j$ and $\tilde{C}_j \gets \tilde{C}_{j-1} \cup C_j$ 
\State Update $\Phi_j$ by $\mathcal{L}$ on $D_j$ (train on current task)
\State $\tilde{M}_j \gets \tilde{M}_{j-1}$
\For{each $r \in R_j$}
    \State pick $L$ samples in $D_j$ and add them into $\tilde{M}_j$
\EndFor
\State $\tilde{R}_j \gets \tilde{R}_{j-1} \cup R_j$
\State Update $\tilde{P}_j$ with new data in $D_j$ (for inference)
\State Update $\Phi_j$ by $\mathcal{L}$ on $\tilde{M}_j$ and $D^*_j$ (train on memory)
\end{algorithmic}
\end{algorithm}

\subsubsection{Testing phase} 
To classify relations during inference, we utilize the \textit{Nearest-Class-Mean} (NCM) classifier, as proposed by \citet{ma-etal-2024-making}. Unlike conventional methods that rely solely on label prototypes, we incorporate both label descriptions and prototypes to improve relation prediction.

Given a sample $x$ with hidden representation $\bm{z}_x$, we define a set of relation prototypes $\{\bm{p}_r\}_{r=1}^n$. Each relation prototype is computed as the mean representation of all support samples associated with that relation:

\begin{equation}
\bm{p}_r = \frac{1}{L} \sum_{i=1}^{L}\bm{z}_i,
\end{equation}

where $L$ is the number of support samples contributing to the prototype.

The relation prediction is determined by computing the cosine similarity between $\bm{z}_x$ and each prototype $\bm{p}_r$ as well as the corresponding label description $\bm{d}_r$. The final prediction $y^*$ is selected based on the highest similarity score:

\begin{equation}
    y^* = \arg\max_r \gamma(\bm{z}_x, \bm{p}_r)
    \label{eq:primary}
\end{equation}



where $\gamma(\cdot, \cdot)$ represents the cosine similarity function. \\
When testing without UR samples to compare with prior work in standard scenarios using only learned relation types, our augmented descriptions effectively represent known relations. However, the defined description for the undetermined relation may not accurately reflect its true labels. Therefore, we use the average description hidden representation instead of a prototype in Equation \ref{eq:primary}.

We also examine the use of Open Information Extraction (OIE) for inference. Specifically, OIE first eliminates entity pairs with no identifiable relationship (No relation - NA), assigning them to the Undetermined Relation (UR) label. For candidate triplets that pass this filtering—categorized as NOTA (None of the Above) or DR (Determined Relation)—a trained language model performs standard matching to classify them into a known relation type or UR. The function utilizing OIE is defined as follows (\ref{fig:infer}):

\begin{align}
    \mathcal{F}(e_i, e_j, x) &= 
    \begin{cases} 
        \text{NA}, \text{if } \text{OIE}(e_i, e_j, x) \rightarrow \text{null} \\
        y^* \text{ (i.e., NOTA or DR)}
    \end{cases}
\end{align}




\begin{table*}[ht]
\centering
\begin{adjustbox}{width=\textwidth}
\begin{tabular}{lllllllll}
\toprule
\multirow{2}{*}{Method} & \multicolumn{8}{c}{Tasks} \\
\cmidrule{2-9}
& \multicolumn{1}{c}{$\mathcal{T}^1$} & \multicolumn{1}{c}{$\mathcal{T}^2$} & \multicolumn{1}{c}{$\mathcal{T}^3$} & \multicolumn{1}{c}{$\mathcal{T}^4$} & \multicolumn{1}{c}{$\mathcal{T}^5$} & \multicolumn{1}{c}{$\mathcal{T}^6$} & \multicolumn{1}{c}{$\mathcal{T}^7$} & \multicolumn{1}{c}{$\mathcal{T}^8$}  \\ 
\toprule
\multicolumn{9}{c}{\textbf{FewRel} \textit{(10-way--5-shot)}} \\
\midrule
SCKD        & $91.92_{\pm 0.80}$ & $79.37_{\pm 4.83}$ & $75.07_{\pm 3.45}$ & $73.72_{\pm 2.15}$ & $69.11_{\pm 2.02}$ & $68.38_{\pm 2.45}$ & $67.18_{\pm 2.10}$ & $65.04_{\pm 5.76}$ \\ 
ConPL & $\mathbf{94.36_{\pm 0.63}}$ & $\underline{84.61_{\pm 3.31}}$ & $\underline{78.41_{\pm 1.93}}$ & $74.16_{\pm 3.41}$ & $72.37_{\pm 2.48}$ & $\underline{71.83_{\pm 3.51}}$ & $\underline{68.45_{\pm 1.67}}$ & $64.46_{\pm 0.71}$ \\
CPL  & {$92.11_{\pm 0.96}$} & $82.94_{\pm 2.89}$ & ${76.64_{\pm 4.50}}$ & $\underline{74.66_{\pm 6.69}}$ & $\underline{73.08_{\pm 6.40}}$ & $69.89_{\pm 5.55}$ & $68.01_{\pm 3.02}$ & $65.29_{\pm 1.38}$ \\ 
CPL\_MI & $\underline{93.15_{\pm 0.61}}$ & {$82.20_{\pm 2.94}$} & {$76.53_{\pm 1.58}$} & {$73.52_{\pm 2.36}$} & {$71.79_{\pm 2.18}$} & {$69.17_{\pm 2.50}$} & {$67.18_{\pm 0.93}$} & \underline{$65.34_{\pm 0.79}$} \\
EDC\textsuperscript{*} & $68.88_{\pm 0.02}$ & {$54.46_{\pm 0.97}$} & {$49.05_{\pm 2.11}$} & {$46.45_{\pm 2.31}$} & {$43.89_{\pm 1.82}$} & {$41.92_{\pm 1.39}$} & {$38.91_{\pm 0.04}$} & {$36.81_{\pm 0.12}$} \\
OFCRE (Ours) & $91.02_{\pm 0.90}$ & $\mathbf{85.36_{\pm 1.91}}$ & $\mathbf{79.83_{\pm 1.68}}$ & $\mathbf{76.46_{\pm 1.60}}$ & $\mathbf{74.69_{\pm 2.73}}$ & $\mathbf{72.08_{\pm 2.01}}$ & $\mathbf{69.60_{\pm 1.44}}$ & $\mathbf{67.62_{\pm 0.95}}${\color{darkgreen}\footnotesize $\uparrow{2.28}$} \\

\toprule
\multicolumn{9}{c}{\textbf{TACRED} \textit{(5-way-5-shot)}} \\
\midrule
SCKD     & \underline{$86.59_{\pm 1.02}$} & $78.91_{\pm 3.97}$ & $70.63_{\pm 4.00}$ & $63.94_{\pm 5.17}$ & $58.41_{\pm 2.27}$ & $57.58_{\pm 4.02}$ & $51.96_{\pm 2.28}$ & $49.43_{\pm 2.09}$ \\
ConPL & {${85.67_{\pm 3.23}}$} & $80.95_{\pm 4.82}$ & $69.85_{\pm 2.05}$ & $61.18_{\pm 3.29}$ & $59.31_{\pm 1.80}$ & $56.02_{\pm 3.76}$ & $54.93_{\pm 2.57}$ & $51.67_{\pm 3.58}$ \\ 
CPL & $86.38_{\pm 0.94}$ & $\underline{81.21_{\pm 3.23}}$ & $\underline{74.09_{\pm 3.32}}$ & $\underline{69.36_{\pm 6.53}}$ & $\underline{63.48_{\pm 4.40}}$ & $\underline{61.36_{\pm 3.77}}$ & $56.09_{\pm 2.88}$ & $\underline{53.81_{\pm 3.11}}$ \\
CPL\_MI  & $\mathbf{86.69_{\pm 1.12}}$ & {$79.87_{\pm 5.20}$} & {$72.34_{\pm 2.90}$} & {$68.85_{\pm 4.18}$} & {$62.61_{\pm 4.95}$} & {$60.05_{\pm 4.66}$} & \underline{$57.34_{\pm 5.21}$} & {$53.59_{\pm 1.78}$} \\ 
EDC\textsuperscript{*}  & $53.29_{\pm 0.02}$ & {$55.18_{\pm 2.31}$} & {$55.53_{\pm 0.17}$} & {$54.77_{\pm 2.44}$} & {$52.66_{\pm 0.56}$} & {$54.10_{\pm 1.87}$} & {$53.47_{\pm 2.42}$} & {$52.93_{\pm 0.04}$} \\
OFCRE (Ours)& $85.23_{\pm 0.39}$ & $\mathbf{82.39_{\pm 2.78}}$ & $\mathbf{77.64_{\pm 3.39}}$ & $\mathbf{74.67_{\pm 3.91}}$ & $\mathbf{71.08_{\pm 5.83}}$ & $\mathbf{70.79_{\pm 3.94}}$ & $\mathbf{68.91_{\pm 2.87}}$ & $\mathbf{67.8_{\pm 1.32}}${\color{darkgreen}\footnotesize  $\uparrow{13.99}$} \\

\toprule
\end{tabular}
\end{adjustbox}
\caption{F1 score (\%) of methods using BERT backbone after training for each task \textbf{without undetermined relation} in dataset. The best results are in \textbf{bold}, while the second highest scores are \underline{underlined}}
\label{table:main_wo_ur}
\end{table*}

\begin{table*}[ht]
\centering
\begin{adjustbox}{width=\textwidth}
\begin{tabular}{lllllllll}
\toprule
\multirow{2}{*}{Method} & \multicolumn{8}{c}{Tasks} \\
\cmidrule{2-9}
& \multicolumn{1}{c}{$\mathcal{T}^1$} & \multicolumn{1}{c}{$\mathcal{T}^2$} & \multicolumn{1}{c}{$\mathcal{T}^3$} & \multicolumn{1}{c}{$\mathcal{T}^4$} & \multicolumn{1}{c}{$\mathcal{T}^5$} & \multicolumn{1}{c}{$\mathcal{T}^6$} & \multicolumn{1}{c}{$\mathcal{T}^7$} & \multicolumn{1}{c}{$\mathcal{T}^8$}  \\ 
\toprule
\multicolumn{9}{c}{\textbf{FewRel} \textit{(10-way--5-shot)}} \\
\midrule
SCKD      & $62.96_{\pm 0.72}$ & $45.05_{\pm 4.93}$ & $36.11_{\pm 3.42}$ & $31.14_{\pm 1.81}$ & $25.76_{\pm 2.04}$ & $25.03_{\pm 2.47}$ & $24.63_{\pm 2.09}$ & $22.99_{\pm 5.60}$ \\ 
ConPL & ${58.28_{\pm 1.06}}$ & $34.65_{\pm 4.37}$ & $32.24_{\pm 2.45}$ & $29.48_{\pm 2.18}$ & $28.66_{\pm 1.70}$ & $28.23_{\pm 3.84}$ & $26.85_{\pm 3.79}$ & $24.49_{\pm 4.68}$ \\
CPL & {$57.00_{\pm 1.22}$} & $28.77_{\pm 2.55}$ & $22.63_{\pm 1.84}$ & $18.86_{\pm 5.17}$ & $15.82_{\pm 5.22}$ & $15.19_{\pm 3.84}$ & $13.44_{\pm 1.23}$ & $13.26_{\pm 1.30}$ \\
CPL\_MI & $58.28_{\pm 1.33}$ & {$34.65_{\pm 3.18}$} & {$32.24_{\pm 3.09}$} & {$29.48_{\pm 3.03}$} & {$28.66_{\pm 3.29}$} & {$28.23_{\pm 3.63}$} & {$26.85_{\pm 1.88}$} & {$24.49_{\pm 1.63}$} \\
EDC\textsuperscript{*} & $37.44_{\pm 0.08}$ & {$31.40_{\pm 1.87}$} & {$28.98_{\pm 1.10}$} & {$26.58_{\pm 2.24}$} & {$24.99_{\pm 1.76}$} & {$24.28_{\pm 1.59}$} & {$22.00_{\pm 0.61}$} & {$20.65_{\pm 0.03}$} \\
OFCRE (Ours) & $\underline{64.98_{\pm 1.31}}$ & $\underline{51.80_{\pm 3.72}}$ & $\underline{46.64_{\pm 2.32}}$ & $\underline{45.11_{\pm 2.34}}$ & $\underline{43.06_{\pm 2.68}}$ & $\underline{40.44_{\pm 1.33}}$ & $\underline{38.92_{\pm 0.84}}$ & $\underline{37.06_{\pm 0.42}}$ \\
OFCRE + OIE (Ours) & $\mathbf{69.23_{\pm 0.53}}$ & $\mathbf{58.23_{\pm 3.58}}$ & $\mathbf{52.94_{\pm 2.84}}$ & $\mathbf{51.17_{\pm 2.89}}$ & $\mathbf{49.31_{\pm 2.68}}$ & $\mathbf{46.57_{\pm 1.34}}$ & $\mathbf{45.00_{\pm 0.88}}$ & $\mathbf{43.11_{\pm 0.87}}${\color{darkgreen}\footnotesize $\uparrow{6.05}$} \\

\toprule
\multicolumn{9}{c}{\textbf{TACRED} \textit{(5-way-5-shot)}} \\
\midrule
SCKD       & {$59.92_{\pm 2.89}$} & $45.93_{\pm 2.29}$ & $27.84_{\pm 4.03}$ & $22.33_{\pm 5.19}$ & $20.74_{\pm 2.09}$ & $17.20_{\pm 3.94}$ & $15.71_{\pm 2.27}$ & $14.76_{\pm 2.04}$ \\
ConPL & {${56.18_{\pm 1.58}}$} & $27.47{\pm 3.61}$ & $25.37_{\pm 2.30}$ & $21.03_{\pm 3.22}$ & $16.82_{\pm 2.08}$ & $17.08_{\pm 3.89}$ & $16.55_{\pm 3.40}$ & $15.38_{\pm 4.34}$ \\ 
CPL & {$57.00_{\pm 0.83}$} & $28.77_{\pm 2.33}$ & $22.63_{\pm 2.10}$ & $18.86_{\pm 5.15}$ & $15.82_{\pm 4.88}$ & $15.19_{\pm 3.51}$ & $13.44_{\pm }1.44$ & $13.26_{\pm }1.76$ \\ 
CPL\_MI & $48.49_{\pm 3.71}$ & {$28.20_{\pm 3.85}$} & {$20.60_{\pm 2.61}$} & {$17.82_{\pm 1.40}$} & {$16.49_{\pm 1.92}$} & {$14.92_{\pm 1.20}$} & {$14.00_{\pm 1.34}$} & {$13.22_{\pm 0.73}$} \\
EDC\textsuperscript{*} & $32.75_{\pm 0.01}$ & {$33.36_{\pm 1.01}$} & {$34.74_{\pm 2.21}$} & {$33.17_{\pm 4.24}$} & {$31.25_{\pm 2.53}$} & {$33.30_{\pm 2.73}$} & {$32.35_{\pm 2.13}$} & {$31.34_{\pm 0.04}$} \\

OFCRE (Ours)& $\underline{65.99_{\pm 0.99}}$ & $\underline{53.08_{\pm 1.71}}$ & $\underline{45.52_{\pm 0.11}}$ & $\underline{41.99_{\pm 5.31}}$ & $\underline{37.79_{\pm 5.64}}$ & $\underline{35.73_{\pm 3.03}}$ & $\underline{33.20_{\pm 2.29}}$ & $\underline{32.15_{\pm 1.48}}$ \\
OFCRE + OIE (Ours)& $\mathbf{67.51_{\pm 0.81}}$ & $\mathbf{59.14_{\pm 1.59}}$ & $\mathbf{52.23_{\pm 0.85}}$ & $\mathbf{48.85_{\pm 5.82}}$ & $\mathbf{43.26_{\pm 5.38}}$ & $\mathbf{41.28_{\pm 3.31}}$ & $\mathbf{38.87_{\pm 2.32}}$ & $\mathbf{37.79_{\pm 1.58}}${\color{darkgreen}\footnotesize  $\uparrow{5.64}$} \\
\toprule
\end{tabular}
\end{adjustbox}
\caption{F1 score (\%) of methods using BERT backbone after training for each task \textbf{with undetermined relation} in dataset. OFCRE + OIE is a test version that utilizes OIE to filter out samples with no relations before passing them to OFCRE. The best results are in \textbf{bold}, while the second highest scores are \underline{underlined}}
\label{table:main_ur}
\end{table*}

\begin{table}[ht]
    \centering
     \resizebox{\columnwidth}{!}{%
    \setlength{\tabcolsep}{1mm}
    \begin{tabular}{lcccc}
    \hline 
    \multirow{2}{*}{Method} & \multicolumn{2}{c}{w/o UR} & \multicolumn{2}{c}{UR}\\
    \cmidrule(lr){2-3} \cmidrule(lr){4-5} 
    & FewRel & TACRED & FewRel & TACRED \\
    \hline \hline \\
    \text{OFCRE (Ours)} & \textbf{67.62} & \textbf{67.80} & \textbf{43.11} & \textbf{37.79}\\

    \quad  \text{w/o $\mathcal{L}_{HSMT}$} & 65.16 &  67.10 & 40.50 & 35.80\\
    
    \quad  \text{w/o $\mathcal{L}_{WMI_{SD}}$} & 63.75 & \underline{67.35} & \underline{42.26} & \underline{36.79}\\
    
    \quad  \text{w/o $\mathcal{L}_{WMI_{SC}}$} & \underline{66.95} & 66.40 & 41.92 & 36.35\\
    \hline
    \end{tabular}
    }
    \caption{Ablation study of loss at task $\mathcal{T}^8$}
    \label{tab:ablation_study}
\end{table}

\section{Experiments}
\subsection{Experiment Setup}

We conduct experiments using the pre-trained language model BERT \citep{devlin-etal-2019-bert} as the backbone for the Few-shot Continual Relation Matching module. We then evaluate our approach against baselines on two widely used benchmarks in the literature of CRE and FCRE: FewRel \citep{han-etal-2018-fewrel} and TACRED \citep{zhang-etal-2017-position}. These datasets are added with numerous samples containing undetermined relations (UR), treated as a new relation type with a corresponding description (see Appendix \ref{app:data}).  

After completing each task, we evaluate the models on the updated $\mathcal{D}^{test}$ using six random seeds and report the mean and standard deviation of accuracy. We benchmark our approach against state-of-the-art baselines under similar settings, including \textbf{SCKD} \citep{wang-etal-2023-serial}, \textbf{ConPL} \citep{DBLP:conf/acl/ChenWS23}, \textbf{CPL} \citep{ma-etal-2024-making}, \textbf{CPL+MI} \citep{tran-etal-2024-preserving}, and \textbf{EDC\textsuperscript{*}}\footnote{EDC\textsuperscript{*} is a modified version of EDC in which the phase OIE prompt has been adjusted to accept a list of entities from our dataset as additional input.} \citep{zhang-soh-2024-extract}.

Since the presence of numerous undetermined relations affects overall relation extraction performance, we evaluate using the \textbf{F1 score} \citep{nguyen2015relation} for determined relations.
\subsection{Experiment results}
 \label{exp_method}

\paragraph{Superior Performance Across FCRE Scenarios} 
Our model consistently achieves superior results across all scenarios, regardless of the inclusion of the UR labels. Particularly, when trained with UR labels and evaluated only on DR labels (Table \ref{table:main_wo_ur}), it exhibits the lowest forgetting rates—23.4\% on Fewrel and 17.42\% on TACRED —while maintaining superior performance across tasks up to the final task, $\mathcal{T}^8$. This highlights its ability to retain knowledge of seen relations. Additionally, as shown in Table \ref{table:main_ur}, our model excels at correctly identifying relation types even in the presence of numerous undetermined relations (URs). It significantly outperforms baselines, which fail when trained and tested solely on UR labels. In the final task, our model surpasses the weakest baseline by 29.85\% on Fewrel and 24.57\% on TACRED. This demonstrates its robustness in leveraging Open Information Extraction descriptions and original description alignment, rather than depending solely on sample-hidden representations.
\paragraph{Our model shows strong potential to support Knowledge Graph Construction in the setting of FCRE} 

When evaluated with only the DR label (Table \ref{table:main_wo_ur}), the latest SOTA EDC performs worse than when UR labels are included (Table \ref{table:main_ur}), following the trend of other methods. In the final task with UR, our model surpasses EDC by 16.41\% on Fewrel and 1.11\% on TACRED. Notably, on Fewrel, where the number of relations increases with each task, EDC struggles with catastrophic forgetting due to its reliance on pretrained models for description search and the lack of suitable matching relations.

In the setting with 
To address this, we introduce an OFCRE variant that leverages Open Information Extraction (OIE) as a filter, ensuring only relevant samples are passed for prediction. This approach mirrors EDC’s extraction and canonicalization phases but consolidates them into a single LLM phase, leading to significantly improved matching and state-of-the-art results: 43.11\% on Fewrel and 37.79\% on Tacred with UR.

\begin{figure}[H]
    \centering
    \includegraphics[width=\columnwidth]{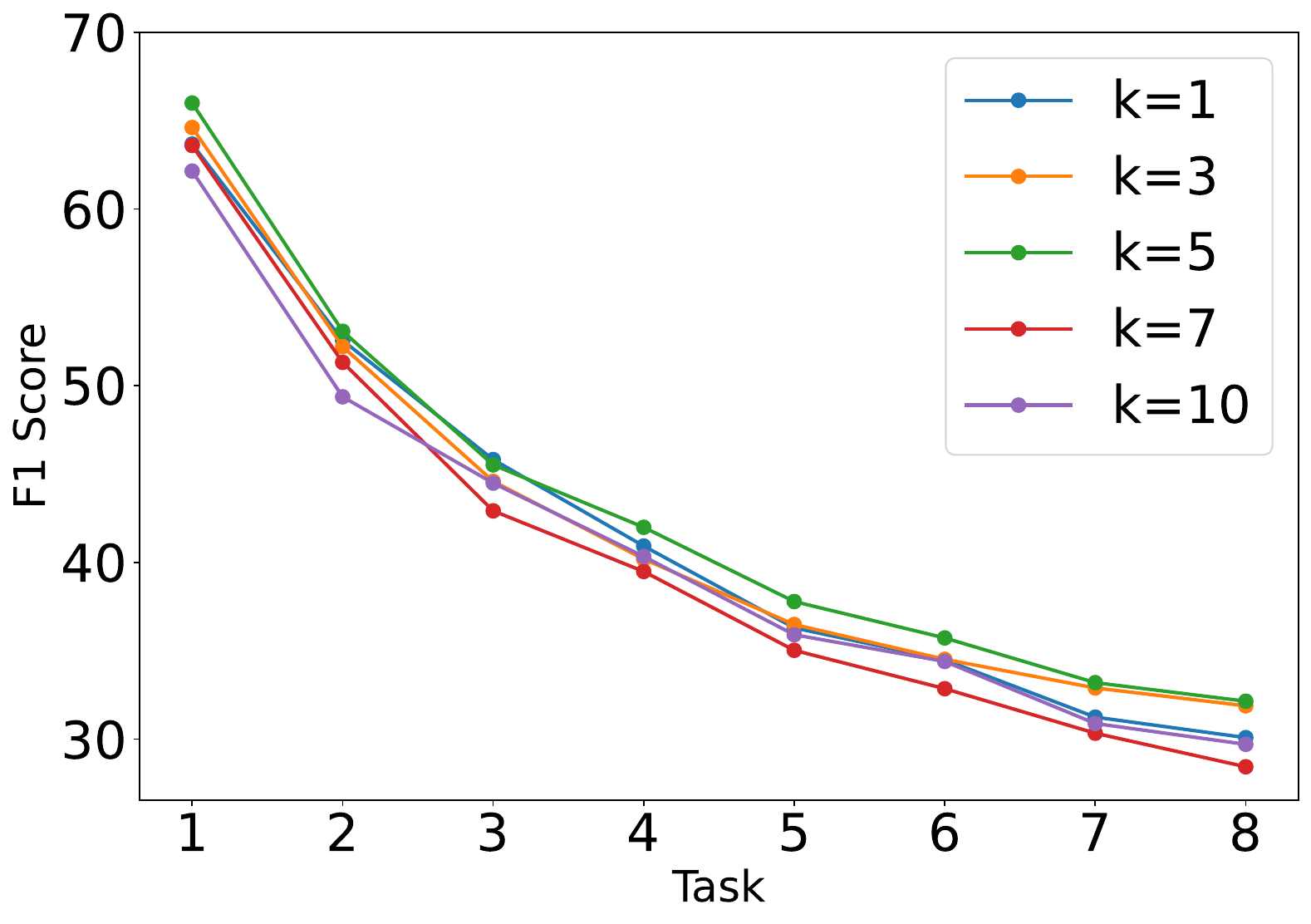}
    \caption{Results with each K augmented description on Tacred with Undetermined Relation}
    \vspace{-2mm}
    \label{fig:kaug1}
\end{figure}

\paragraph{The effects of using augmented descriptions}
Figure \ref{fig:kaug1} shows that \( K = 5 \) optimally balances diversity and noise, achieving peak accuracy (\SI{74.67}{\%} on \textbf{TACRED} $\mathcal{T}^4$). Smaller \( K \) (\( \leq 3 \)) limits contextual diversity, while larger \( K \) (\( =7 \)) introduces noise, reducing performance by \(\sim\)2--3\%. Higher \( K \) benefits \textbf{TACRED} due to noisy text. Augmented descriptions enhance undetermined relation (UR) detection (\SI{12.4}{\%} gain over \( K = 1 \)), though computational cost and redundancy remain challenges.

\paragraph{Ablation Study}
In Table \ref{tab:ablation_study}, we present the performance variations when removing the core losses $\mathcal{L}_{HSMT}$, $\mathcal{L}_{WMI_{SD}}$, and $\mathcal{L}_{WMI_{SC}}$. The results demonstrate the contribution of each loss function to the overall performance. Notably, incorporating both $\mathcal{L}_{WMI_{SD}}$ and $\mathcal{L}_{WMI_{SC}}$ losses, rather than using only one, improves performance by up to 4\%. This emphasizes the significance of both the original and candidate descriptions, even when numerous undetermined relations need to be extracted.

\section{Conclusion}
We present a novel approach to \textit{Few-shot Continual Relation Extraction} that integrates \textit{Open Information Extraction (OIE)} for addressing undetermined relations. By incorporating leveraging OIE and structured descriptions, our method effectively improves performance of FCRE models. In addition, we are also the first work that explores and elevates \textit{Knowledge Graph Construction (KGC)} in the setting of Continual Learning. Experiments confirm state-of-the-art performance, demonstrating the effectiveness of our proposed method.
\section{Limitation}
The dataset labeled with UR (undetermined relations) may, in fact, contain instances that align with predefined relation types, though the original annotators did not indicate this. As we cannot be sure that every instance is a true relation, we continue to assign them to the UR category and test using the same data and settings for all methods to ensure fairness. Note that further data verification is beyond the scope of this work. Moreover, training and testing with a large number of undetermined relations is computationally expensive and time-consuming. Therefore, optimizing this process can be considered as future work to improve efficiency.
\newpage
\bibliography{anthology,custom}
\bibliographystyle{acl_natbib}


\clearpage  
\appendix

\section*{Appendix}  
\addcontentsline{toc}{section}{Appendix}  

\renewcommand{\thesection}{\Alph{section}}  

\section{Experimental Details}
    
\subsection{Few-shot Continual Relation Extraction (FCRE)}
\subsubsection{Benchmark datasets}
\label{app:data}
Our experiments for the FCRE scenario utilize two benchmark datasets:

\begin{itemize}
    \item \textbf{FewRel} \citep{han-etal-2018-fewrel}: This dataset comprises 100 relations with 70,000 samples. Following \citet{DBLP:conf/acl/QinJ22}, we employ a configuration of 80 relations, partitioned into 8 tasks, each containing 10 relations (10-way). The initial task, $\mathcal{T}^1$, includes 100 samples per relation, while subsequent tasks are structured as few-shot tasks under 5-shot settings.
\begin{table}[h]
    \centering
    \resizebox{\columnwidth}{!}{%
    \begin{tabular}{lcc}
        \toprule
        & \textbf{Train} & \textbf{Test} \\
        \midrule
        Total samples of DR & 1350 & 8000 \\
        Total samples of UR & 7431 & 43175 \\
        Average entities per sample & 4.21 & 4.20 \\
        \bottomrule
    \end{tabular}
    }
    \caption{Fewrel Dataset with UR Statistics}
    \label{tab:first1}
\end{table}

    \item \textbf{TACRED} \citep{zhang-etal-2017-position}: This dataset encompasses 42 relations with 106,264 samples extracted from Newswire and Web documents. Consistent with \citep{DBLP:conf/acl/QinJ22}, we exclude instances labeled as ``no\_relation'' and distribute the remaining 41 relations across 8 tasks. The first task, $\mathcal{T}^1$, comprises 6 relations with 100 samples each, while subsequent tasks involve 5 relations (5-way) in 5-shot configurations.
\begin{table}[h]
    \centering
    \resizebox{\columnwidth}{!}{%
    \begin{tabular}{lcc}
        \toprule
        & \textbf{Train} & \textbf{Test} \\
        \midrule
        Total samples of DR & 775 & 2122 \\
        Total samples of UR & 5173 & 15152 \\
        Average entities per sample & 4.43 & 4.61 \\
        \bottomrule
    \end{tabular}
    }
    \caption{Tacred Dataset with UR Statistics}
    \label{tab:second2}
\end{table}
    
\end{itemize}

In addition to adding samples and their corresponding entities with an undetermined relation, we also incorporate this description into the training data.

Definition of \textbf{Undetermined relation}: This relation is used when the relationship between entities is either not applicable or unknown. It serves as a default category when no other relation type clearly applies or when there is insufficient information to determine the relationship.

\subsubsection{Example Sample in Dataset}
\label{app:example}
\textbf{Example Sample:}  
Spearhafoc was succeeded by William the Norman and was the last Bishop of London of English descent for an extended period, likely until Roger Niger's appointment in 1228.

\textbf{Determined Relation:}  
The entities \textit{Spearhafoc} and \textit{Bishop of London} are determined to have the relation type: \textit{Location of Formation}, as per the dataset. \\
\textbf{Undetermined Relation} is one of two types below:
\begin{itemize}
\item \textbf{None of the Above (NOTA)}: The relation between \textit{Roger Niger} and \textit{Bishop of London} is classified as \textit{"Appointed Location"}.

\item \textbf{No Relation (NA):}  
The entities \textit{Roger Niger} and \textit{Spearhafoc} do not share any directly applicable relation.
\end{itemize}
\subsection{Baselines}
\label{sec:appendix_baseline1}

This study evaluates our approach against state-of-the-art methods in FCRE and FCED. The selected baselines are as follows:

\subsubsection{FCRE Baselines}
\begin{itemize}
    \item \textbf{SCKD} \citep{DBLP:conf/acl/WangWH23} introduces a structured knowledge distillation approach aimed at retaining information from past tasks. This method incorporates contrastive learning along with pseudo samples to improve the differentiation capability of relation representations.
    
    \item \textbf{ConPL} \citep{DBLP:conf/acl/ChenWS23} consists of three key components: a prototype-based classification module, a memory-enhanced mechanism, and a consistency learning module that helps maintain distributional stability while reducing forgetting. Moreover, ConPL leverages prompt learning to refine representation learning and applies focal loss to minimize ambiguity among similar classes.

    \item \textbf{CPL} \citep{DBLP:conf/coling/MaHL024} proposes a framework that employs prompts to enhance generalization across categories and adopts margin-based contrastive learning to manage difficult samples, effectively addressing catastrophic forgetting and overfitting. Additionally, CPL integrates a memory augmentation technique using ChatGPT to generate diverse samples, further alleviating overfitting in low-resource FCRE settings.

    To perform the ablation study presented in Table ...

    \item \textbf{CPL+MI} \citep{tran-etal-2024-preserving} (Mutual Information Maximization) serves as an enhancement to existing baselines by utilizing the often-overlooked language model heads to retain prior knowledge from pre-trained backbones and improve representation learning. This is accomplished by maximizing the mutual information between the latent representations from the language model head branch and the primary classifier branch.
\end{itemize}

\subsubsection{Open Information Extraction baselines}
EDC \citep{zhang-soh-2024-extract}, Extract-Define-Canonicalize, is a novel framework designed for knowledge graph construction (KGC) using large language models (LLMs). KGC is the task of creating knowledge graphs, which are structured representations of knowledge that organize interconnected information through graph structures, with entities and relations represented as nodes and edges. The EDC framework addresses the challenges of using LLMs for KGC, particularly in scenarios with large or unavailable schemas.

The key idea behind EDC is to \textbf{break down KGC into three phases}:

\begin{enumerate}
    \item \textbf{Open Information Extraction (OIE)}: 
    This phase involves extracting entity-relation triplets from the input text without adhering to a pre-defined schema. Large Language Models (LLMs) are used to identify and extract these triplets. For example, given the following text:
        "Alan Shepard was born on Nov 18, 1923 and selected by NASA in 1959. He was a member of the Apollo 14 crew."

    The extracted triplets might be:
    \begin{itemize}
        \item ("Alan Shepard", "bornOn", "Nov 18, 1923")
        \item ("Alan Shepard", "participatedIn", "Apollo 14")
    \end{itemize}
    
    \item \textbf{Schema Definition}: 
    In this phase, LLMs generate natural language definitions for each relation type identified in the extraction phase. For the example above, definitions for ``bornOn'' and ``participatedIn'' would be generated.

    \item \textbf{Schema Canonicalization}: 
    This phase refines the open knowledge graph into a canonical form by eliminating redundancies and ambiguities. This is done either through \textit{target alignment} (with an existing target schema) or \textit{self-canonicalization} (without a target schema). In target alignment, the system identifies the most closely related components within the target schema for each element, and LLMs assess the feasibility of each potential transformation. For instance,  LLMs will replace ``participatedIn'' in the retrieved closest schema (``mission'', ``season'', etc.) to ``mission''. In self-canonicalization, the system consolidates semantically similar schema components, standardizing them to a singular representation.
\end{enumerate}

To further improve performance, the EDC framework can be iteratively refined with a \textbf{Schema Retriever}. The Schema Retriever is a trained model that retrieves schema components relevant to the input text, akin to retrieval-augmented generation. This process involves constructing a ``hint'' for the extraction phase, which includes candidate entities and relations extracted in previous iterations.

The benefits of EDC include its \textbf{flexibility, performance, and ability to handle large schemas or situations where no pre-defined schema is available}. Experiments have demonstrated that EDC can extract higher-quality knowledge graphs compared to state-of-the-art methods. EDC is also more general compared to existing canonicalization methods because it works whether a target schema is provided or not. Instead of using static external sources like WordNet, EDC utilizes contextual and semantically-rich side information generated by LLMs. Furthermore, by allowing the LLMs to verify if a transformation can be performed, EDC alleviates the over-generalization issue faced by previous methods.

There are several limitations that could be addressed in future works. These include incorporating an entity de-duplication mechanism, improving the components of EDC (such as the schema retriever), testing smaller open-source models' performance on the other tasks, and reducing the cost of EDC.

\subsection{Training Configurations}
This section outlines the optimal hyperparameter configurations utilized across our experimental framework. Through comprehensive Grid Search optimization, we identified the optimal values for loss factors $\alpha_{x}$, $\alpha_{xd}$, and $\alpha_{xc}$ by exploring the range $[0.5, 1.0, 2.0, 3.0]$. Table \ref{tab:hyperparameters_cpl1} details the specific parameter settings for each model variant.


\begin{table}[ht]
 \resizebox{\columnwidth}{!}{%
    \begin{tabular}{lc}
        \hline
        \textbf{Hyperparameter} & \textbf{Value} \\
        \hline
        Number of training epochs & 10 \\
        Memory training epochs & 10 \\
        Learning rate & $1 \times 10^{-5}$ \\
        \hline
        Encoder output dimension & 768 \\
        BERT input maximum sequence length & 256 \\
        \hline
        $\alpha_x$ & 1.0 \\
        $\alpha_{xd}$ & 2.0 \\
        $\alpha_{xc}$ & 2.0 \\
        \hline
    \end{tabular}
    }
    \caption{Hyperparameter configuration for OFCRE}
    \label{tab:hyperparameters_cpl1}
\end{table}
General process of inference in OFCRE using OIE:
\begin{figure}[ht]
    \centering
    \includegraphics[width=\columnwidth]{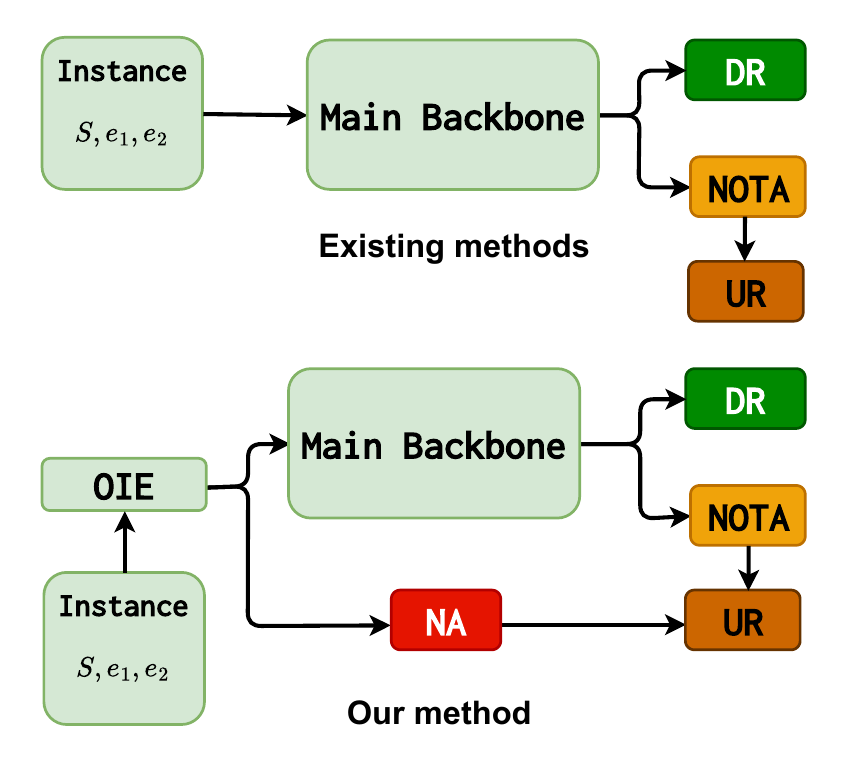}
    \caption{In case the labels of the pair of entities are not learned yet, we can know whether the pair of entities has a meaningful relationship or not.}
    \label{fig:infer}
\end{figure}
\begin{table*}[ht]
\centering
\begin{adjustbox}{width=\textwidth}
\begin{tabular}{lllllllll}
\toprule
\multirow{2}{*}{Method} & \multicolumn{8}{c}{Tasks} \\
\cmidrule{2-9}
& $\mathcal{T}^1$ & $\mathcal{T}^2$ & $\mathcal{T}^3$ & $\mathcal{T}^4$ & $\mathcal{T}^5$ & $\mathcal{T}^6$ & $\mathcal{T}^7$ & $\mathcal{T}^8$  \\ 
\toprule
\multicolumn{9}{c}{\textbf{FewRel} \textit{(10-way--5-shot)}} \\
\midrule
\method & $91.02_{\pm 0.90}$ & $\mathbf{85.36_{\pm 1.91}}$ & $\mathbf{79.83_{\pm 1.68}}$ & $\mathbf{76.46_{\pm 1.60}}$ & $\mathbf{74.69_{\pm 2.73}}$ & $\mathbf{72.08_{\pm 2.01}}$ & $\mathbf{69.60_{\pm 1.44}}$ & $\mathbf{67.62_{\pm 0.95}}$ \\

\quad w/o $\mathcal{L}_{HSMT}$ & ${91.23_{\pm 0.50}}$ & $\underline{84.73_{\pm 2.65}}$ & $\underline{79.30_{\pm 2.15}}$ & $\underline{76.28_{\pm 1.96}}$ & \underline{$73.68_{\pm 2.62}$} & \underline{$71.07_{\pm 2.43}$} & $\underline{68.92_{\pm 1.20}}$ & ${65.16_{\pm 0.41}}$ \\

\quad w/o $\mathcal{L}_{WMI_{SD}}$ & $\mathbf{92.88_{\pm 0.87}}$ & $84.09_{\pm 1.77}$ & $78.4_{\pm 1.83}$ & $74.29_{\pm 1.62}$ & $71.25_{\pm 3.25}$ & $68.3_{\pm 3.07}$ & $66.02_{\pm 1.67}$ & $63.75_{\pm 0.86}$ \\

\quad w/o $\mathcal{L}_{WMI_{SC}}$ & \underline{$91.57_{\pm 0.78}$} & {$84.19_{\pm 1.97}$} & {$79.24_{\pm 2.61}$} & {$75.17_{\pm 2.33}$} & ${73.59_{\pm 3.14}}$ & ${70.65_{\pm 2.39}}$ & {$68.57_{\pm 1.40}$} & \underline{$66.95_{\pm 0.25}$} \\
\toprule
\multicolumn{9}{c}{\textbf{TACRED} \textit{(5-way-5-shot)}} \\
\midrule

\method & $85.23_{\pm 0.39}$ & $\mathbf{82.39_{\pm 2.78}}$ & $\mathbf{77.64_{\pm 3.39}}$ & $\mathbf{74.67_{\pm 3.91}}$ & ${71.08_{\pm 5.83}}$ & $\mathbf{70.79_{\pm 3.94}}$ & $\mathbf{68.91_{\pm 2.87}}$ & $\mathbf{67.8_{\pm 1.32}}$ \\

\quad w/o $\mathcal{L}_{HSMT}$ & $\mathbf{85.8_{\pm 0.64}}$ & ${81.22_{\pm 3.58}}$ & \underline{$77.07_{\pm 4.96}$} & \underline{$74.47_{\pm 3.56}$} & $\mathbf{71.95_{\pm 5.13}}$ & {$69.31_{\pm 3.28}$} & $\underline{68.58_{\pm 1.41}}$ & $67.10_{\pm 2.33}$ \\

\quad w/o $\mathcal{L}_{WMI_{SD}}$ & $85.49_{\pm 1.15}$ & $\underline{82.12_{\pm 2.85}}$ & $76.47_{\pm 3.39}$ & $73.98_{\pm 2.65}$ & $70.86_{\pm 3.04}$ & \underline{$70.51_{\pm 1.54}$} & $67.51_{\pm 2.35}$ & $\underline{67.35_{\pm 2.04}}$ \\

\quad w/o $\mathcal{L}_{WMI_{SC}}$ &$\underline{85.52_{\pm 0.77}}$ & $81.50_{\pm 4.15}$ & $75.32_{\pm 3.66}$ & $73.71_{\pm 3.78}$ & $\underline{71.11_{\pm 5.42}}$ & $69.40_{\pm 4.37}$ & {$67.11_{\pm 3.56}$} & {$66.40_{\pm 1.86}$} \\

\bottomrule

\end{tabular}
\end{adjustbox}
\caption{Ablation study (\%) of loss functions for our model tested \textbf{without undetermined relation}. The best results are in \textbf{bold}, while the second highest scores are \underline{underlined}}
\label{table:ablation1}
\end{table*}

\begin{table*}[ht]
\centering
\begin{adjustbox}{width=\textwidth}
\begin{tabular}{lllllllll}
\toprule
\multirow{2}{*}{Method} & \multicolumn{8}{c}{Tasks} \\
\cmidrule{2-9}
& $\mathcal{T}^1$ & $\mathcal{T}^2$ & $\mathcal{T}^3$ & $\mathcal{T}^4$ & $\mathcal{T}^5$ & $\mathcal{T}^6$ & $\mathcal{T}^7$ & $\mathcal{T}^8$  \\ 
\toprule
\multicolumn{9}{c}{\textbf{FewRel} \textit{(10-way--5-shot)}} \\
\midrule
\method & $\underline{64.98_{\pm 1.31}}$ & $\mathbf{51.80_{\pm 3.72}}$ & $\mathbf{46.64_{\pm 2.32}}$ & $\mathbf{45.11_{\pm 2.34}}$ & $\mathbf{43.06_{\pm 2.68}}$ & $\mathbf{40.44_{\pm 1.33}}$ & $\mathbf{38.92_{\pm 0.84}}$ & $\mathbf{37.06_{\pm 0.42}}$ \\

\quad w/o $\mathcal{L}_{HSMT}$ & ${61.75_{\pm 0.62}}$ & ${49.8_{\pm 4.05}}$ & ${44.18_{\pm 2.56}}$ & ${40.52_{\pm 3.07}}$ & {$40.19_{\pm 2.71}$} & {$37.39_{\pm 1.42}$} & ${35.74_{\pm 1.48}}$ & ${33.39_{\pm 0.70}}$ \\

\quad w/o $\mathcal{L}_{WMI_{SD}}$ & $\mathbf{65.52_{\pm 0.60}}$ & $50.2_{\pm 3.62}$ & $\underline{45.46_{\pm 1.67}}$ & $\underline{44.73_{\pm 2.68}}$ & $\underline{42.30_{\pm 2.76}}$ & $39.03_{\pm 1.71}$ & $36.65_{\pm 1.49}$ & $\underline{35.77_{\pm 0.98}}$ \\

\quad w/o $\mathcal{L}_{WMI_{SC}}$ & $63.28_{\pm 0.43}$ & \underline{$50.72_{\pm 3.95}$} & {$44.11_{\pm 2.25}$} & {$43.53_{\pm 2.60}$} & ${41.13_{\pm 2.71}}$ & $\underline{39.88_{\pm 2.17}}$ & \underline{$37.92_{\pm 0.85}$} & {$35.30_{\pm 1.40}$} \\
\toprule
\multicolumn{9}{c}{\textbf{TACRED} \textit{(5-way-5-shot)}} \\
\midrule

\method & $\mathbf{65.99_{\pm 0.99}}$ & $\underline{53.08_{\pm 1.71}}$ & $\mathbf{45.52_{\pm 0.11}}$ & $\mathbf{41.99_{\pm 5.31}}$ & $\mathbf{37.79_{\pm 5.64}}$ & $\mathbf{35.73_{\pm 3.03}}$ & $\mathbf{33.20_{\pm 2.29}}$ & $\mathbf{32.15_{\pm 1.48}}$ \\

\quad w/o $\mathcal{L}_{HSMT}$ & ${61.42_{\pm 2.71}}$ & ${49.29_{\pm 3.31}}$ & {$42.81_{\pm 4.18}$} & {$36.34_{\pm 3.34}$} & {$33.02_{\pm 3.03}$} & {$30.30_{\pm 1.25}$} & $29.11_{\pm 0.85}$ & $27.20_{\pm 1.36}$ \\

\quad w/o $\mathcal{L}_{WMI_{SD}}$ & $62.66_{\pm 0.57}$ & $52.55_{\pm 2.97}$ & $43.41_{\pm 2.07}$ & $\underline{41.26_{\pm 3.31}}$ & $36.79_{\pm 3.53}$ & $\underline{34.85_{\pm 3.07}}$ & $32.46_{\pm 1.12}$ & $31.52_{\pm 1.09}$ \\

\quad w/o $\mathcal{L}_{WMI_{SC}}$ &$\underline{64.34_{\pm 0.60}}$ & $\mathbf{53.92_{\pm 1.69}}$ & $\underline{44.84_{\pm 1.66}}$ & $40.41_{\pm 4.01}$ & $\underline{37.15_{\pm 5.09}}$ & $34.29_{\pm 2.97}$ & \underline{$32.88_{\pm 2.08}$} & \underline{$31.54_{\pm 1.39}$} \\

\bottomrule

\end{tabular}
\end{adjustbox}
\caption{Ablation study (\%) of loss functions for our model tested \textbf{with undetermined relation}. The best results are in \textbf{bold}, while the second highest scores are \underline{underlined}}
\label{table:ablation2}
\end{table*}

\begin{table*}[ht]
\centering
\begin{adjustbox}{width=\textwidth}
\begin{tabular}{lllllllll}
\toprule
\multirow{2}{*}{Method} & \multicolumn{8}{c}{Tasks} \\
\cmidrule{2-9}
& $\mathcal{T}^1$ & $\mathcal{T}^2$ & $\mathcal{T}^3$ & $\mathcal{T}^4$ & $\mathcal{T}^5$ & $\mathcal{T}^6$ & $\mathcal{T}^7$ & $\mathcal{T}^8$  \\ 
\toprule
\multicolumn{9}{c}{\textbf{FewRel} \textit{(10-way--5-shot)}} \\
\midrule
\method + OIE & $\underline{69.23_{\pm 0.53}}$ & $\mathbf{58.23_{\pm 3.58}}$ & $\mathbf{52.94_{\pm 2.84}}$ & $\mathbf{51.17_{\pm 2.89}}$ & $\mathbf{49.31_{\pm 2.68}}$ & $\mathbf{46.57_{\pm 1.34}}$ & $\mathbf{45.00_{\pm 0.88}}$ & $\mathbf{43.11_{\pm 0.87}}$ \\

\quad w/o $\mathcal{L}_{HSMT}$ & ${67.33_{\pm 0.69}}$ & ${57.07_{\pm 3.78}}$ & ${51.40_{\pm 2.68}}$ & ${47.97_{\pm 2.85}}$ & {$47.22_{\pm 2.65}$} & {$44.31_{\pm 1.48}$} & ${42.7_{\pm 1.28}}$ & ${40.5_{\pm 0.53}}$ \\

\quad w/o $\mathcal{L}_{WMI_{SD}}$ & {$\mathbf{69.43_{\pm 0.48}}$} & $\underline{57.77_{\pm 3.46}}$ & $\underline{52.90_{\pm 1.71}}$ & $\underline{50.03_{\pm 2.64}}$ & $\underline{48.56_{\pm 2.65}}$ & $\underline{45.42_{\pm 1.82}}$ & $\underline{44.03_{\pm 1.48}}$ & $\underline{42.26_{\pm 0.87}}$ \\

\quad w/o $\mathcal{L}_{WMI_{SC}}$ & $68.07_{\pm 0.18}$ & {$56.94_{\pm 3.64}$} & {$50.45_{\pm 2.39}$} & {$49.94_{\pm 2.54}$} & ${48.52_{\pm 2.79}}$ & ${45.31_{\pm 2.22}}$ & {$43.46_{\pm 0.90}$} & {$41.92_{\pm 1.12}$} \\
\toprule
\multicolumn{9}{c}{\textbf{TACRED} \textit{(5-way-5-shot)}} \\
\midrule

\method + OIE & $\mathbf{67.51_{\pm 0.81}}$ & $\mathbf{59.14_{\pm 1.59}}$ & $\mathbf{52.23_{\pm 0.85}}$ & $\mathbf{48.85_{\pm 5.82}}$ & $\mathbf{43.26_{\pm 5.38}}$ & $\mathbf{41.28_{\pm 3.31}}$ & $\mathbf{38.87_{\pm 2.32}}$ & $\mathbf{37.79_{\pm 1.58}}$ \\

\quad w/o $\mathcal{L}_{HSMT}$ & ${67.26_{\pm 1.17}}$ & ${57.09_{\pm 2.33}}$ & {$50.84_{\pm 3.47}$} & {$45.03_{\pm 4.23}$} & {$41.56_{\pm 4.54}$} & {$39.08_{\pm 1.96}$} & $37.55_{\pm 0.95}$ & $35.80_{\pm 1.77}$ \\

\quad w/o $\mathcal{L}_{WMI_{SD}}$ & $67.05_{\pm 0.69}$ & $57.45_{\pm 2.56}$ & $\underline{51.70_{\pm 1.70}}$ & $\underline{47.89_{\pm 2.96}}$ & $\underline{42.93_{\pm 4.17}}$  & $\underline{42.26_{\pm 2.87}}$ & $\underline{39.24_{\pm 1.70}}$ & $\underline{36.79_{\pm 1.17}}$\\

\quad w/o $\mathcal{L}_{WMI_{SC}}$ &$\underline{67.36_{\pm 0.36}}$ & $\underline{58.07_{\pm 2.00}}$ & $50.85_{\pm 1.68}$ & $46.73_{\pm 4.16}$ & $41.10_{\pm 5.33}$ & $41.03_{\pm 2.92}$ & {$37.54_{\pm 2.62}$} & {$36.35_{\pm 1.61}$} \\

\bottomrule

\end{tabular}
\end{adjustbox}
\caption{Ablation study (\%) of loss functions for our model with OIE tested \textbf{with undetermined relation}. The best results are in \textbf{bold}, while the second highest scores are \underline{underlined}}
\label{table:ablation3}
\end{table*}

\section{Ablation study}
The ablation study highlights the critical role of each loss component in \textbf{OFCRE}'s performance. We further analyze the contributions of the \textbf{Hard Soft Margin Triplet Loss ($\mathcal{L}_{HSMT}$)}, \textbf{Weighted Mutual Information Loss for raw descriptions ($\mathcal{L}_{WMI_{SD}}$)}, and \textbf{Weighted Mutual Information Loss for candidate descriptions ($\mathcal{L}_{WMI_{SC}}$)} across tasks and datasets.

\begin{itemize}
    \item \textbf{Impact of $\mathcal{L}_{HSMT}$}: 
    Removing $\mathcal{L}_{HSMT}$ leads to a $\sim$2--4\% drop in accuracy on both datasets. For instance, on \textbf{TACRED} (Table~\ref{table:ablation1}), Task $\mathcal{T}^8$ performance declines from \SI{67.8}{\%} to \SI{65.16}{\%} without UR. This loss enforces margin-based separation between hardest positive and negative pairs, critical for distinguishing semantically similar relations.
    
    \item \textbf{Role of $\mathcal{L}_{WMI_{SD}}$}: 
    Excluding $\mathcal{L}_{WMI_{SD}}$ results in gradual performance decay, with a $\sim$3--5\% decline by $\mathcal{T}^8$ on \textbf{FewRel} (Table~\ref{table:ablation1} and Table~\ref{table:ablation2}). This loss aligns sample representations with raw relation descriptions (e.g., \textit{"headquarters location: administrative center"}), stabilizing knowledge retention.
    
    \item \textbf{Significance of $\mathcal{L}_{WMI_{SC}}$}: 
    $\mathcal{L}_{WMI_{SC}}$ is critical for handling \textbf{undetermined relations (UR)}. On \textbf{TACRED} with UR (Table~\ref{table:ablation2}), removing it causes a $\sim$5\% drop in $\mathcal{T}^8$ (\SI{32.15}{\%} $\rightarrow$ \SI{31.54}{\%}). This loss leverages OIE-generated candidate descriptions (e.g., \textit{"was born in"} for \textit{"person place of birth"}) to generalize to unseen relations.
\end{itemize}

\begin{itemize}
    \item \textbf{FewRel}: Dominated by $\mathcal{L}_{HSMT}$ and $\mathcal{L}_{WMI_{SD}}$, as predefined relations are structured. UR detection relies less on OIE candidates here.
    \item \textbf{TACRED}: $\mathcal{L}_{WMI_{SC}}$ plays a stronger role due to noisy, real-world text. UR labels often require contextual OIE descriptions for accurate classification.
\end{itemize}


The joint optimization of $\mathcal{L}_{HSMT}$, $\mathcal{L}_{WMI_{SD}}$, and $\mathcal{L}_{WMI_{SC}}$ ensures a balance between \textit{discriminative power}, \textit{knowledge retention}, and \textit{adaptability to unseen relations}. Future work could explore:
\begin{itemize}
    \item Dynamic loss weighting (e.g., increasing $\mathcal{L}_{WMI_{SC}}$'s weight in later tasks).
    \item Reinforcement learning to prioritize underperforming components during training.
\end{itemize}
\label{sec:appendix_ablation1}


\begin{table*}[ht]
    \centering
    \begin{mdframed}[linewidth=1pt] 
        \textbf{Prompt for generating an expanded relation description with examples from original description} \\
        \hrulefill
        You are an experienced data scientist working on a relation extraction task. \\
        Your objective is to take a given relation and its brief description and produce a more detailed explanation. Additionally, you should generate three diverse sentence examples demonstrating the relation in use. \\
        The relation is: headquarters location \\
        The description is: location where an organization's central administration is based. \\
        Please provide 2 distinct (relation description, examples) pairs. \\
        Your response:
    \end{mdframed}

    \vspace{5pt} 

    \begin{mdframed}[linewidth=1pt] 
        \textbf{Output:} \\
        \hrulefill
        This relation indicates the primary location where an organization, corporation, or institution's central administrative functions are managed. The headquarters serves as the main hub for decision-making, strategic planning, and coordination of business operations.\\
        Examples: \\
        - The headquarters of Apple Inc. is located in Cupertino, California, where key corporate decisions and product development take place. \\
        - The United Nations Headquarters is based in New York City, serving as the central meeting place for international diplomatic activities. \\
        - Toyota's global headquarters is situated in Toyota City, Japan, overseeing its worldwide automobile manufacturing and business operations. \\

        \vspace{5pt} 

        This relation describes the official site of an organization's main offices, which serves as the administrative center and often houses executives and key departments.\\
        Examples: \\
        - Google's headquarters, known as the Googleplex, is in Mountain View, California, hosting thousands of employees working on technology and innovation. \\
        - The European Central Bank has its headquarters in Frankfurt, Germany, where major financial policies for the Eurozone are formulated. \\
        - The headquarters of Amazon is located in Seattle, Washington, guiding the company's global e-commerce and cloud computing strategies.
    \end{mdframed}

    \caption{Example of generating an expanded relation description with examples from original description}
    \label{tab:relation_description_prompt1}
\end{table*}

\section{Prompt and Output Example}
\label{app:prompt}
\begin{table*}[ht]
    \centering
    \begin{mdframed}[linewidth=1pt] 
        \textbf{Prompt for Open Information Extraction} \\
        \hrulefill
Given a piece of text, two entities subject, object (not ordered) and corresponding relation type between two entities, extract the relation trigger in the form of [Subject, Relation, Object] from it. If there is not any relation, relation is null.
Here are some examples: \\
\\
Example 1: \\
Text: "he passed away on saturday ."\\
Subject, Object entities(not ordered): "he", "saturday" \\
Complete triplets: ["he", "passed away on", "saturday"] \\
\\
Example 2: \\
Text: "as a substantial shareholder in cnac's subsidiary air china, cathay pacific said late monday it would give serious consideration to joining cnac and form a strategic partnership with china eastern." \\
Subject, Object entities(not ordered): "cnac", "cathay pacific" \\
Complete triplets: ["cathay pacific", "a substantial shareholder", "cnac"] \\
\\
Now it's your turn! Please extract the relation from the following text: \\
Text: "Elon Musk founded SpaceX in 2002." \\
Subject, Object (not ordered): "Elon Musk", "SpaceX" \\
Complete triplets: 
    \end{mdframed}

    \vspace{5pt} 

    \begin{mdframed}[linewidth=1pt] 
        \textbf{Output:} \\
        \hrulefill
      ["Elon Musk", "founded", "SpaceX"] \\
        
        
    \end{mdframed}

\caption{Prompt used for Open Information Extraction. During training, it generates a candidate triplet, whereas in testing, it returns only triplet with identified relationship. If no relationship exists (NA), the relation is set to null, e.g., ["Elon Musk", null, "SpaceX"].}
    \label{tab:prompt_stability2}
\end{table*}

\begin{table*}[ht]
    \centering
    \begin{mdframed}[linewidth=1pt] 
        \textbf{Prompt for Relation Definition and Augmentation from Open Information Extraction} \\
        \hrulefill
Define the relationship in a relational triplet extracted from a given text and provide 3 sentence examples of the relationship.
You must generate 2 diverse samples of (relation definition, example) pairs for the relationship. \\

Example 1: \\
Text: "Albert Einstein was born in Germany in 1879." \\
Triplet: ["Albert Einstein", "was born in", "Germany"] \\
Relation type: "person place of birth" \\
Definitions and examples of "was born in": \\

\textbf{Sample 1:}  
\begin{quote}
\{
    "definition": "The relationship between a person and the place where they were born.", \\
    "examples": [ \\
    \quad "Isaac Newton was born in England in 1643.", \\
    \quad "Marie Curie was born in Warsaw, Poland.", \\
    \quad "Leonardo da Vinci was born in Vinci, Italy." \\
    ]
\}
\end{quote}

Now it's your turn! Please define the relationship in the following relational triplet: \\
Text: "Elon Musk founded SpaceX in 2002." \\
Triplet: ["Elon Musk", "founded", "SpaceX"] \\
Relation type: "organization founder" \\
Definitions and examples of "founded":
    \end{mdframed}

    \vspace{5pt} 

    \begin{mdframed}[linewidth=1pt] 
        \textbf{Output:} \\
        \hrulefill
        
        \textbf{Sample 1:} \\
        \begin{quote}
        \{
            "definition": "The relationship between a person and an organization they established.", \\
            "examples": [ \\
            \quad "Bill Gates founded Microsoft in 1975.", \\
            \quad "Steve Jobs founded Apple in 1976.", \\
            \quad "Mark Zuckerberg founded Facebook while studying at Harvard." \\
            ]
        \}
        \end{quote}

        \vspace{5pt} 

        \textbf{Sample 2:} \\
        \begin{quote}
        \{
            "definition": "The connection between an individual and the company or organization they initiated, often as its creator or co-founder.", \\
            "examples": [ \\
            \quad "Larry Page and Sergey Brin founded Google in 1998.", \\
            \quad "Jeff Bezos founded Amazon in 1994.", \\
            \quad "Jack Ma founded Alibaba in 1999." \\
            ]
        \}
        \end{quote}
    \end{mdframed}

    \caption{Candidate Description Definition and Augmentation}
    \label{tab:prompt_stability3}
\end{table*}


\end{document}